\documentclass[10pt,twocolumn,letterpaper]{article}

\usepackage{cvpr}
\usepackage{times}
\usepackage{epsfig}
\usepackage{graphicx}
\usepackage{amsmath}
\usepackage{amssymb}
\usepackage{bm}
\usepackage{bbm}
\usepackage{bbold}
\usepackage{subcaption}
\usepackage{multirow}
\usepackage[font=footnotesize,skip=0pt]{caption}

\cvprfinalcopy % *** Uncomment this line for the final submission

\def\cvprPaperID{****} % *** Enter the CVPR Paper ID here

\ifcvprfinal\fi
\begin{document}

%%%%%%%%% TITLE
\title{SCOUT: Self-aware Discriminant Counterfactual Explanations}

\author{Pei Wang \qquad Nuno Vasconcelos\\
Department of Electrical and Computer Engineering\\
University of California, San Diego\\
{\tt\small \{pew062,nuno\}@ucsd.edu}
% For a paper whose authors are all at the same institution,
% omit the following lines up until the closing ``}''.
% Additional authors and addresses can be added with ``\and'',
% just like the second author.
% To save space, use either the email address or home page, not both
% \and
% Second Author\\
% Institution2\\
% First line of institution2 address\\
% {\tt\small secondauthor@i2.org}
}

\maketitle
%\thispagestyle{empty}

%%%%%%%%% ABSTRACT
\begin{abstract}
  The problem of counterfactual visual explanations is considered.
  A new family of discriminant explanations is introduced.
  These produce heatmaps that attribute high
  scores to image regions informative of a classifier prediction but not
  of a counter class. They connect attributive explanations, which are
  based on a single heat map, to counterfactual explanations, which
  account for both predicted class and counter class. The latter are shown to
  be computable by combination of two discriminant explanations,
  with reversed class pairs. It is argued that
  self-awareness, namely the ability to produce classification confidence
  scores, is important for the computation of discriminant explanations,
  which seek to identify regions where it is easy to discriminate between
  prediction and counter class. This suggests the computation of discriminant
  explanations by the combination of three attribution maps.
  The resulting counterfactual explanations are optimization free
  and thus much faster than previous methods. To address the difficulty
  of their evaluation, a proxy task and set
  of quantitative metrics are also proposed. Experiments under this protocol
  show that the proposed counterfactual explanations outperform the state of
  the art while achieving much higher speeds, for popular
  networks. In a human-learning machine teaching experiment,
  they are also shown to improve mean student accuracy from
  chance level to $95\%$.
\end{abstract}

%%%%%%%%% BODY TEXT
\section{Introduction}

Deep learning (DL) systems are
difficult to deploy in specialized domains, such as medical diagnosis or
biology, requiring very fine-grained distinctions
between visual features unnoticeable to
the untrained eye. Two main difficulties arise. The first is the
black-box nature of DL. When high-stakes decisions are involved, e.g. a
tumor diagnosis, the system users, e.g. physicians, require a justification
for its predictions. The second
is the large data labeling requirements of DL.
Since supervised training is usually needed for optimal
classification, modern networks are trained with large
datasets, manually annotated on
Amazon MTurk. However, because MTurk annotators lack
domain expertise, the
approach does not scale to specialized domains.

\begin{figure}[t]
  \centering
  \includegraphics[width=0.5\textwidth]{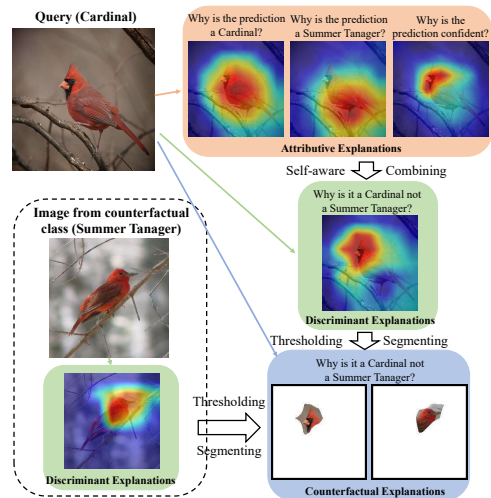}
  \caption{Given a query image (Cardinal) and a counterfactual class
    (Summer Tanager), discriminant explanations bridge the gap between
    attributions and counterfactual explanations. This enables a fast
    optimization-free computation of the latter.}
  \label{fig:teaser}
\end{figure}

Both problems can be addressed by explainable AI (XAI) techniques, which
complement network predictions with human-understandable explanations.
These can both circumvent the black-box nature of DL and
enable the design of machine teaching systems that provide feedback
to annotators when they make mistakes~\cite{zhou2018an}.
%Although there have been some attempts at natural
%language~\cite{hendricks2016generating,kanehira2018learning},
%most algorithms produce visual explanations. Among these,
In computer vision, the dominant XAI paradigm is {\it attribution,\/} which
consists of computing a heatmap
% encoding how the network prediction can be attributed to each image
of how strongly each image
pixel~\cite{simonyan2013deep,bach2015pixel,shrikumar2017learning,ancona2017unified} or
region~\cite{zhou2016learning,selvaraju2017grad} contributes to a
network prediction. For example, when
asked ``why is this a truck?'' an
attributive system would answer or visualize something like
``because it has wheels, a hood, seats,
a steering wheel, a
flatbed, head and tail lights, and rearview mirrors.''

While useful to a naive user, this explanation is less useful to
an expert in the domain. The latter is likely to be interested
in more {\it precise\/} feedback, asking instead the question
``Why is it not a car?'' The answer ``because it has a
flatbed. If it did not have a flatbed it would be a car,''
is known as a {\it counterfactual\/} or {\it contrastive
explanation\/} \cite{wachter2017counterfactual,dhurandhar2018explanations,miller2018contrastive}. Such explanations are more desirable in expert
domains. When faced with a prediction of lesion $A$, a doctor
would naturally ask ``why $A$ but not $B$?'' The same question
would be posed by a student that incorrectly assigned an
image to class
$B$ upon receiving feedback that it belongs to class $A$.
By supporting a specific query with respect to a {\it counterfactual\/}
class ($B$), these
explanations allow expert users to zero-in on a specific
ambiguity between two classes, which they already {\it know\/} to be
plausible prediction outcomes. Unlike attributions, counterfactual
explanations scale naturally with user expertise.
As the latter increases, the class and counterfactual class simply become more
{\it fine-grained.\/}
% This type of scalability is not immediate for
% attributive explanations, since the coarser and most obvious attributions
% for the prediction (``it has tail'') tend to drown
% out the more subtle ones (``the whiskers curl downward'').

In computer vision, counterfactual explanations have only
recently received attention. They are usually implemented as
``correct class is $A$. Class $B$ would require changing the
image as follows,'' where ``as follows'' is some visual
transformations. Possible transformations include
image perturbations~\cite{dhurandhar2018explanations}, synthesis~\cite{wachter2017counterfactual}
or the exhaustive search of a large feature pool, to find replacement
features that map the image from class $A$ to $B$~\cite{pmlr-v97-goyal19a}.
However, image perturbations and synthesis frequently leave
the space of natural images only working on simple non-expert domains, and feature search is too complex
for interactive applications.
%such as machine teaching or medical diagnosis systems.

In this work, a new procedure is proposed to generate {\it Self-aware
  disCriminant cOUnterfactual explanaTions\/} (SCOUT). We show that
counterfactual explanations can be
much more efficiently generated by a combination of attributive
explanations and self-awareness mechanisms, which quantify the
confidence of the predictions of a DL system.
For this, we start by introducing {\it discriminant explanations\/} that,
as shown in Figure~\ref{fig:teaser}, connect attributive to counterfactual
explanations. Like attributive explanations, they consist of a single
heatmap. This, however, is an attribution map for
the {\it discrimination\/} of classes $A$ and $B$, attributing
high scores to image regions that are informative of $A$ but not of $B$.
In this sense, discriminant explanations are similar to counterfactual
explanations and  more precise than attributive
explanations (see Figure~\ref{fig:teaser}). A {\it counterfactual explanation\/} can then be produced by the computation of two
discriminant explanations, with the roles of $A$ and $B$ reversed.

We next consider how to compute discriminant
explanations and argue for the importance of self-awareness.
A system is self-aware if it can {\it quantify the
confidence\/} with which it classifies an image. This is generally
true for DL systems, which complement a class prediction with
an estimate of the posterior class distribution, from which
a confidence score can be derived~\cite{geifman2017selective,wang2017idk}. The attribution map of this score is
an indicator of the image regions where the classification is easy.
This fits nicely in the discriminant explanation framework, where
the goal is to find the spatial regions predictive of class A but
unpredictive of class B. It leads to the {\it definition of
discriminant explanations\/} as image regions that simultaneously: 1)
have high attribution for class $A$, 2) have low attribution
for class $B$, and 3) are classified with high confidence.
It follows that, as shown in Figure~\ref{fig:teaser}, discriminant
explanations can be computed by combination of three attribution
maps. This, in turn, shows that counterfactual explanations can be seen as a
{\it generalization\/} of attributive explanations and computed by a
{\it combination\/} of attribution~\cite{simonyan2013deep,bach2015pixel,shrikumar2017learning,sundararajan2017axiomatic,ancona2017unified} and confidence
prediction methods~\cite{geifman2017selective,wang2017idk,Wang_2018_ECCV}
that is much more efficient to compute than previous methods.

Beyond explanations, a significant challenge to XAI is the lack
of explanation ground truth for performance evaluation. Besides user-based
evaluations~\cite{pmlr-v97-goyal19a},
whose results are difficult to replicate, we propose a quantitative
metric based on a proxy localization task. To the best of our
knowledge, this is the first proposal for semantically quantitative evaluation of counterfactual
visual explanations independently of human experiments. Compared to
the latter, the proposed proxy evaluation is substantially easier to
replicate.  This evaluation shows that SCOUT {\it both\/} outperforms the state of
the art \cite{pmlr-v97-goyal19a}
and is $50\times$ to $1000\times$ faster for popular networks.
This is quite important for applications such as machine teaching,
where explanation algorithms should operate in real-time, and
ideally in low-complexity platforms such as mobile devices.

Overall, the paper makes five contributions. First, a new family of
discriminant explanations, which are substantially more precise than
attributive explanations. Second, the use of self-awareness to
improve the accuracy of attributive explanations. Third, the
derivation of counterfactual explanations by combination of
discriminant explanations, making them more efficient to compute.
Fourth, a new experimental protocol for quantitative evaluation
of counterfactual explanations. Fifth, experimental results using both this
protocol and machine teaching experiments, showing that the
proposed SCOUT outperforms previous methods
and is substantially faster.

% It should be noting that speed is extremely important and can not be ignored. Low computation can guarantee machine teaching algorithm embedded in mobile device for real-time applications. Also, fast response time during feedback could optimize the user experience.

% We summarize our contributions as follows:
% \begin{itemize}
%     \item performance improvement;
%     \item Speed, much more efficient;
%     \item Our regions can have more reception field unit, but their can only find one unit, if they want to find more, it really really time-consuming;
%     \item An evaluation metrics is proposed.
% \end{itemize}

% Speed is important, for machine teaching, the feedback time long should be short. If wait for a long time, the learner will complain. Also, fast speed is a potential pre-requirement for two-stage detection.

% High recall rate is an advantage for region proposal.

\section{Related work}

In this section we review the literature on explanations, self-awareness,
and machine teaching.

\noindent{\bf Explanations:}
Two main approaches to explainable AI (XAI) have emerged in
computer vision. Natural language (NL) systems attempt to produce a
textual explanation understandable to humans~\cite{hendricks2018generating,
Hendricks_2018_ECCV,rathi2019generating}. Since
image to text translation is still a difficult problem, full blown
NL explanations tend to target specific applications, like self
driving~\cite{deruyttere2019talk2car}. More robust systems tend to use a limited vocabulary,
e.g. a set of image attributes~\cite{Hendricks_2018_ECCV,hendricks2018generating}. For example,
\cite{Hendricks_2018_ECCV} proposed counterfactual NL
image descriptions and \cite{hendricks2018generating} produces
counterfactual explanations by extracting noun
phrases from the counter-class, which are filtered with an evidence checker.
Since phrases are defined by attributes, this boils down to detecting
presence/absence of attributes in the query image. These methods require
a priori definition of a vocabulary (e.g. attributes), training data
for each vocabulary term, and
training of the classifier to produce this side information. Due to
these difficulties, most explanation methods
rely instead on visualizations. While the
ideas proposed in this work could be extended to NL
systems, we consider only visual explanations.

\noindent{\bf Attributive explanations:} The most popular approach to
visual explanations is to rely on
attributions~\cite{bach2015pixel,shrikumar2017learning,
sundararajan2017axiomatic}. These methods produce
a heatmap that encodes how much the classifier prediction can be
attributed to each pixel or image region. Many attribution
functions have been
proposed~\cite{simonyan2013deep,bach2015pixel,shrikumar2017learning,
sundararajan2017axiomatic,ancona2017unified}. The most
popular framework is to compute some variant of the gradient of the
classifier prediction with respect a chosen layer of the network
and then backproject to the input~\cite{selvaraju2017grad,zhou2016learning}. These techniques
tend to work well when the object of the predicted class is immersed in
a large background (as in object detection), but are less useful when
the image contains the object alone (as in recognition). In this setting,
the most suitable for the close inspection required in
expert domains, the heat map frequently covers the whole object.
This is illustrated in Figure~\ref{fig:teaser}. Counterfactual explanations,
which involve differences with respect to a counterfactual class, tend not
to suffer  from this problem.

\noindent{\bf Counterfactual explanations:} Given an image of class $A$ and
a counterfactual class $B$, counterfactual explanations (also known
as contrastive~\cite{dhurandhar2018explanations})
produce an image transformation that elicits the classification as
$B$~\cite{van2019interpretable,wachter2017counterfactual,liu2019generative,zintgraf2017visualizing}. The simplest example are adversarial attacks~\cite{dhurandhar2018explanations,van2019interpretable,zhou2018an},
which optimize perturbations to map an image of class $A$ into class $B$. However, adversarial perturbations usually push the
perturbed image outside the boundaries of the space of natural images. Generative methods have been proposed to address
this problem, computing large perturbations that generate realistic
images~\cite{liu2019generative,luss2019generating}. This is guaranteed by the introduction
of regularization
constraints, auto-encoders, or GANs~\cite{goodfellow2014generative}.
However, because realistic images are difficult to synthesize,
these approaches have only been applied to simple MNIST or CelebA~\cite{liu2015faceattributes}
style datasets, not expert domains.
A more plausible alternative is to exhaustively search the space of
features extracted from a large collection of images, to find replacement
features that map the image from class $A$ to $B$~\cite{pmlr-v97-goyal19a}.
While this has been shown to perform well on fine-grained
datasets, exhaustive search is too complex for interactive applications.

\noindent{\bf Evaluation:} The performance of explanation algorithms
is frequently only illustrated by the display of visualizations. In some cases,
explanations are evaluated quantitatively with recourse to human experiments.
This involves the design of a system to elicit user feedback on how
trustful a deep learning system is~\cite{selvaraju2017grad,pmlr-v97-goyal19a,dhurandhar2018explanations,pei2018deliberative}
or evaluate if explanations improve user performance on some
tasks~\cite{pmlr-v97-goyal19a}. While we present results of this type, they
have several limitations: it can be difficult to replicate system
design, conclusions can be affected by the users that
participate in the experiments, and the experiments can be cumbersome to
both set up and perform. In result, the experimental results are rarely
replicable or even comparable. This hampers the scalable evaluation of
algorithms. In this work, we introduce a quantitative protocol for the
evaluation of counterfactual explanations, which overcomes these problems.

\noindent{\bf Self-awareness:} Self-aware systems are systems with some
abilities to measure their limitations or predict failures.
This includes topics such as out-of-distribution
detection~\cite{hendrycks17baseline,liang2017enhancing,
  devries2018learning,lee2017training,li2020background}
or open set recognition~\cite{scheirer2012toward,bendale2016towards}, where
classifiers are trained to reject non-sensical images, adversarial attacks,
or images from classes on which they were not trained. All these problems
require the classifier to produce a confidence score for image rejection.
The most popular solution is to guarantee that the
posterior class distribution is uniform, or has high entropy, outside
the space covered by training
images~\cite{lee2017training,hendrycks2018deep}. This, however,
is not sufficient
for counterfactual explanations, which require more precise confidence
scores explicitly addressing class $A$ or $B$. In this sense, the latter
are more closely related to realistic classification~\cite{Wang_2018_ECCV},
where a classifier must identify and reject examples that it deems
too difficult to classify.

\noindent{\bf Machine teaching:} Machine teaching systems \cite{zhou2018an}
are usually designed to teach some tasks to human learners, e.g. image labeling.
These systems usually leverage a model of student learning to optimize teaching
performance~\cite{singla2014near,basu2013teaching,le2004teaching,
piech2015deep}. Counterfactual explanations are naturally suited for
machine teaching, because they provide feedback on why a mistake
(the choice of the counterfactual class $B$) was made.
While the goal of this work is not to design a full blown machine teaching
system, we investigate if counterfactual explanations can improve
human labeling performance. This follows the protocol introduced
by~\cite{pmlr-v97-goyal19a}, which highlights matching bounding boxes on paired
images (what part of $A$ should be replaced by what part of $B$) to provide
feedback to students. Besides improved labeling performance, the proposed
explanations are orders of magnitude faster than the exhaustive
search of~\cite{pmlr-v97-goyal19a}.

\section{Discriminant Counterfactual Explanations}

In this section, we briefly review the main ideas behind previous
explanation approaches and introduce the proposed explanation technique.

\noindent{\bf Counterfactual explanations:} Consider a recognition problem, mapping images $\mathbf{x} \in \mathcal{X}$
into classes $y \in \mathcal{Y} = \{1, \ldots, C\}$.
Images are classified by an object recognition system $\mathcal{H}:
\mathcal{X} \rightarrow \mathcal{Y}$ of the form
\begin{equation}
  y^* = \arg\max_y h_y({\bf x}),
\end{equation}
where ${\bf h}({\bf x}): {\cal X} \rightarrow [0,1]^C$ is a C-dimensional
probability distribution with $\sum_{y=1}^C h_y({\bf x}) = 1$, usually
computed by a convolutional neural network (CNN). The classifier
is learned on a training set $\mathcal{D}$ of $N$ i.i.d.
samples $\mathcal{D} = \{ (\mathbf{x}_i, y_i)\}^N_{i=1}$, where
$y_i \in \mathcal{Y}$ is the label of image $\mathbf{x}_i \in \mathcal{X}$,
and its performance evaluated on a test set
$\mathcal{T} = \{ (\mathbf{x}_j, y_j)\}^M_{j=1}$.
Given an image $\mathbf{x}$, for which the classifier predicts
class $y^*$, counterfactual explanations answer the question
of why the image does not belong to a counterfactual class (also denoted
counter class) $y^c \neq y^*$,
chosen by the user who receives the explanation.

\noindent{\bf Visual explanations:} Counterfactual explanations for
 vision systems are usually based on visualizations.
Two possibilities exist. The first is to explicitly transform the
image $\mathbf{x}$ into an image $\mathbf{x}^c$ of class $y^c$, by
replacing some of its pixel values. The transformation can consist
of applying an image perturbation akin to
those used in adversarial attacks~\cite{dhurandhar2018explanations},
or replacing regions of $\mathbf{x}$ by regions of some images in the
counter class $y^c$~\cite{pmlr-v97-goyal19a}. Due to the difficulties of realistic image synthesis, these methods
are only feasible when $\mathbf{x}$ is relatively simple, e.g. an MNIST
digit.
%In general, requiring the solution of image synthesis
%as a precursor to the explanation problem does not
%appear sensible to us.

A more plausible alternative is to use an already available
image $\mathbf{x}^c$ from class $y^c$ and highlight the differences
between $\mathbf{x}$ and $\mathbf{x}^c$. \cite{pmlr-v97-goyal19a}
proposed to do this by displaying matched bounding boxes on the two images,
and showed that explanation performance is nearly independent of the
choice of $\mathbf{x}^c$, i.e. it suffices to use a random
image $\mathbf{x}^c$ from class $y^c$. We adopt a similar strategy in this work.
For these approaches, the explanation consists of
\begin{equation}
  {\cal C}(\mathbf{x},y^*, y^c, \mathbf{x}^c)
  =  (\mathbf{c}^*(\mathbf{x}), \mathbf{c}^c(\mathbf{x}^c)),
  \label{eq:cfe}
\end{equation}
where $\mathbf{c}^*(\mathbf{x})$ and $\mathbf{c}^c(\mathbf{x}^c)$
are {\it counterfactual heatmaps\/} for images
$\mathbf{x}$ and $\mathbf{x}^c$, respectively, from which region
segments $\mathbf{r}^*(\mathbf{x})$ and $\mathbf{r}^c(\mathbf{x}^c)$ can
be obtained, usually by thresholding.
The question is how to compute these heatmaps. \cite{pmlr-v97-goyal19a}
proposed to search by exhaustively matching all combinations
of features in $\mathbf{x}$ and $\mathbf{x}^c$, which is 
expensive. In this work, we propose a much simpler and more
effective procedure that leverages a large literature on
{\it attributive\/} explanations.

\noindent{\bf Attributive explanations:} Attributive explanations are
a family of explanations based on the attribution of the
prediction $y^*$ to regions of
$\mathbf{x}$~\cite{simonyan2013deep,bach2015pixel,shrikumar2017learning,
  sundararajan2017axiomatic,ancona2017unified}. They are usually produced
by applying an attribution function to a tensor
of activations ${\bf F} \in \mathbb{R}^{W \times H \times D}$ of spatial
dimensions $W \times H$ and $D$ channels, extracted at any layer of a
deep network. While many attribution functions have been proposed,
they are usually some variant of the gradient of $h_{y^*}(\mathbf{x})$ with
respect to ${\bf F}$. This results in an {\it attribution map\/}
$a_{i,j}(\mathbf{x})$ whose amplitude encodes the attribution of the prediction
to each entry $i,j$ along the spatial dimensions of ${\bf F}$.
Attributive explanations produce heat maps of the form
\begin{equation}
  \mathcal{A}(\mathbf{x}, y^*) = \mathbf{a}(h_{y^*}(\mathbf{x}))
\end{equation}
for some attribution function $\mathbf{a}(.)$. Two examples of attributive heatmaps of an image of a "Cardinal," with respect to
predictions "Cardinal" and "Summer Tanager," are shown in the top row of Figure~\ref{fig:teaser}.

\noindent{\bf Discriminant explanations:} In this work, we propose
a new class of explanations, which is denoted as {\it discriminant\/} and defined as
\begin{equation}
  {\cal D}(\mathbf{x},y^*, y^c)
  =  \mathbf{d}(h_{y^*}(\mathbf{x}),h_{y^c}(\mathbf{x})),
  \label{eq:de}
\end{equation}
which have commonalities with both attributive and
counterfactual explanations. Like counterfactual explanations,
they consider both the prediction $y^*$ and
a counterfactual class $y^c$. Like attributive explanations,
they compute a {\it single attribution map\/} through $\mathbf{d}(.,.)$.
The difference is that this map {\it attributes the discrimination
between the prediction $y^*$ and counter
$y^c$ class\/} to regions of $\mathbf{x}$. While $\mathbf{a}(h_{y^*}(\mathbf{x}))$
assigns large attribution to pixels that are strongly informative of
class $y^*$, $\mathbf{d}(h_{y^*}(\mathbf{x}),h_{y^c}(\mathbf{x}))$ does
the same to pixels that are {\it strongly informative\/} of
class $y^*$ but {\it uninformative\/} of class $y^c$.

Discriminant explanations can be used to compute counterfactual
explanations by implementing~(\ref{eq:cfe}) with
\begin{equation}
  {\cal C}(\mathbf{x}, y^*, y^c, \mathbf{x}^c)
  =  ({\cal D}(\mathbf{x},y^*, y^c),
  {\cal D}(\mathbf{x}^c, y^c, y^*)).
\end{equation}
The first map identifies the regions of $\mathbf{x}$ that
are informative of the predicted class but not the counter class
while the second identifies the regions of $\mathbf{x}^c$
 informative of the counter class but not of the predicted class.
Altogether, the explanation shows that
the regions highlighted in the two images are matched:
the region of the first image depicts features that {\it only\/} appear
in the predicted class while that of the second depicts features
that {\it only\/} appear in the counterfactual class. Figure~\ref{fig:teaser}
illustrates the construction of a counterfactual explanation with two
discriminant explanations.

\begin{figure*}[t]
  \centering
  \includegraphics[width=0.9\textwidth]{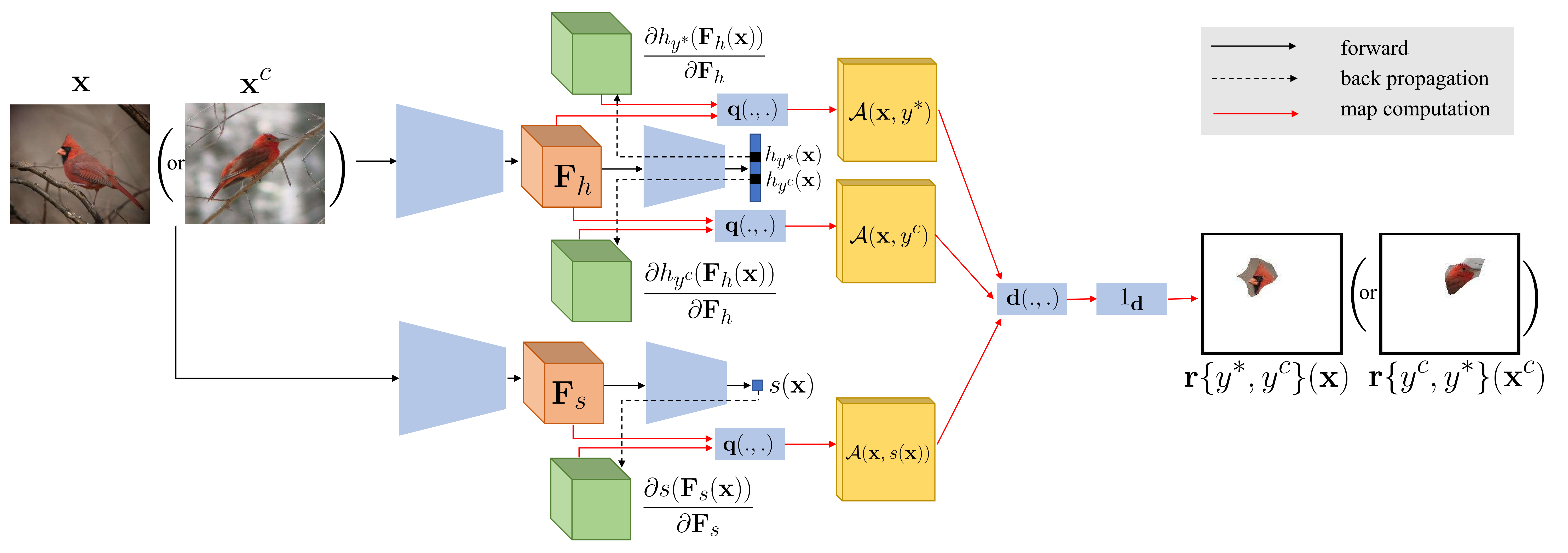}
  \caption{Discriminant explanation architecture ($\mathbf{x}$: Cardinal,
    $\mathbf{x}^c$: Summer Tanager.). Feature activations $\mathbf{F}_h$ and
  $\mathbf{F}_s$ are computed for some layers of the classifier (upper branch) and confidence
  predictor (lower branch), respectively. Attributions for prediction $h_{y^*}$, counter
  class $h_{y^c}$, and confidence score $s$ are computed by attribution
  functions $q(.,.)$ according to (\ref{eq:gradient}) and then combined
  with~(\ref{equ:self_dis}) to obtain the discriminant map. Counterfactual explanations are obtained
  by reversing the roles of $\mathbf{x}$ and $\mathbf{x}^c$ and thresholding
  the discriminant heat maps.}
  \label{architecture}
\end{figure*}

\noindent{\bf Self-awareness:}
% In this work, we seek an approach to
% counterfactual explanations that does not require computationally
% intensive operations, such explicit optimization of perturbations or
% exhaustive search over a large feature pool.
Discriminant maps could be computed by
combining attributive explanations with respect to the predicted
and counter class. Assuming that binary
ground truth segmentation maps $s^*_{i,j}$ and $s^c_{i,j}$ are available
for the attributions of the predicted and counter classes,
respectively, this could be done with the
segmentation map $s_{i,j} = s^*_{i,j} . (1- s^c_{i,j}).$
This map would identify image regions attributable to
the predicted class $y^*$ but not the counter class $y^c$.
In practice,
 segmentation maps are not available and can only be estimated
from attribution maps $a^*_{i,j}$ and $a^c_{i,j}$.
While this could work well when the two classes are
very different, it is not likely to work when they are similar.
This is because, as shown in Figure~\ref{fig:teaser}, attribution maps usually
cover substantial parts of the object. When the two classes differ only in
small parts or details, they lack the precision to
allow the identification of the associated regions. This is critical
for expert domains, where users are likely to ask questions involving
very similar classes.

Addressing this problem requires some ways to sharpen attribution maps.
In this work, we advocate for the use of self-awareness. We assume that
the classifier produces a {\it confidence score} $s({\bf x}) \in [0,1]$,
which encodes the strength of its belief that the image belongs to
the predicted class. Regions that clearly belong to the
predicted class $y^*$ render a score close to $1$ while regions that
clearly do not render a score close to $0$.
This score is {\it self-referential} if generated by the
classifier itself and {\it not self-referential} if generated by a
separate network. The discriminant maps of~(\ref{eq:de}) are then
implemented as
\begin{equation}
  \mathbf{d}(h_{y^*}(\mathbf{x}),h_{y^c}(\mathbf{x})) =
  \mathbf{a}(h_{y^*}(\mathbf{x})) \cdot
  \overline{\mathbf{a}}(h_{y^c}(\mathbf{x})) \cdot
  \mathbf{a}(s(\mathbf{x}))
  \label{equ:self_dis}
\end{equation}
where $\overline{\mathbf{a}}(.)$ is the complement of
$\mathbf{a}(.)$, i.e.
\begin{equation}
  \overline{\mathbf{a}}_{i,j} = \max_{i,j} \mathbf{a}_{i,j} - \mathbf{a}_{i,j}.
\end{equation}

The discriminant map $\mathbf{d}$ is large only at locations $(i,j)$
that contribute strongly to the prediction of class $y^*$ but little to that
of class $y^c$, {\it and where the discrimination between the two classes is
easy, i.e. the classifier is confident.\/} This, in turn, implies that location $(i,j)$ is strongly specific to
class $y^*$ but non specific to class $y^c$, which is the essence of the
counterfactual explanation. Figure~\ref{fig:teaser} shows how the
self-awareness attribution map is usually much sharper than the other two maps.

\noindent{\bf Segmentations:} For discriminant explanations, the
discriminant map of $\mathbf{x}$ is thresholded to obtain the
segmentation mask
\begin{equation}
\mathbf{r}\{y^*,y^c\}({\mathbf x}) =
\mathbbm{1}_{\mathbf{d}(h_{y^*}(\mathbf{x}),h_{y^c}(\mathbf{x}))>T},
\label{eq:segdisc}
\end{equation}
where $\mathbbm{1}_{\cal S}$ is the indicator function of set $\cal S$ and
$T$ a threshold. For counterfactual explanations, segmentation masks
are also generated for $\mathbf{x}^c$, using
\begin{equation}
  \mathbf{r}\{y^c,y^*\}({\mathbf x}^c) =
  \mathbbm{1}_{\mathbf{d}(h_{y^c}(\mathbf{x}^c),h_{y^*}(\mathbf{x}^c))>T}.
  \label{eq:segcounter}
\end{equation}

\noindent{\bf Attribution maps:} The attribution maps of (\ref{equ:self_dis})
can be computed with any attribution function $\mathbf{a}(.)$ in the literature~\cite{sundararajan2017axiomatic,shrikumar2017learning,bach2015pixel}. In our implementation, we use the gradient-based
function of~\cite{shrikumar2016not}.
This calculates the dot-product of the partial
derivatives of the prediction $p$ with respect to the activations
$\mathbf{F}(\mathbf{x})$ of a CNN layer and the activations, i.e.
% \begin{equation}
%   a_{i,j}(h_p) = q\left({\bf F},  \frac{\partial h_p({\bf F})}
%   {\partial {\bf f}_{i,j}}\right) = \frac{\partial h_p({\bf F})}{\partial {\bf f}_{i,j}}
%   \cdot {\bf f}_{i,j}
%   \label{eq:gradient}
% \end{equation}
\begin{equation}
  a_{i,j}(h_p) =
  q\left({\bf f}_{i,j},  \frac{\partial h_p({\bf F})}
  {\partial {\bf f}_{i,j}}\right) =
  \left\langle \frac{\partial h_p({\bf F})}{\partial {\bf f}_{i,j}} ,
    {\bf f}_{i,j} \right\rangle,
  \label{eq:gradient}
\end{equation}
where we omit the dependency on $\mathbf{x}$ for simplicity.

\noindent{\bf Confidence scores:}
Like attribution maps, many existing confidence or hardness scores can be
leveraged. We considered three scores of different characteristics.
The {\it softmax score\/}~\cite{geifman2017selective} is
the largest class posterior probability
\begin{equation}
  s^{s}({\bf x}) = \max_y h_y({\bf x}).
  \label{eq:so}
\end{equation}
It is computed  by adding a max pooling layer to the
network output. The  {\it certainty score\/} is the complement
of the normalized entropy of the softmax
distribution~\cite{wang2017idk},
\begin{equation}
  s^{c}({\bf x}) = 1 + \frac{1}{\log C}\sum_y h_y({\bf x}) \log h_y({\bf x}).
  \label{eq:ce}
\end{equation}
Its computation requires an additional layer of $\log$ non-linearities
and average pooling. These two scores are self-referential. We also
consider the non-self-referential {\it easiness  score\/} of~\cite{Wang_2018_ECCV},
\begin{equation}
  s^{e}({\bf x}) = 1- s^{hp}({\bf x})
  \label{eq:ea}
\end{equation}
where $s^{hp}({\bf x})$ is computed by an external hardness
predictor $\mathcal{S}$, which is
jointly trained with the classifier. $\mathcal{S}$
is implemented with a network $s^{hp}({\bf x}): {\cal X} \rightarrow [0,1]$
whose output is a sigmoid unit.

\noindent{\bf Network implementation:}  Figure \ref{architecture} shows a network implementation of (\ref{equ:self_dis}). Given a query image $\mathbf{x}$ of class $y^*$, a user-chosen counter class $y^c \neq  y^*$, a predictor $h_{y}(\mathbf{x})$, and a confidence predictor $s(\mathbf{x})$ are used to produce the explanation. Note that $s(\mathbf{x})$ can share weights with $h_{y}(\mathbf{x})$ (self-referential) or be  separate (non-self-referential). $\mathbf{x}$ is forwarded through the network, generating activation tensors $\mathbf{F}_h(\mathbf{x})$, $\mathbf{F}_s(\mathbf{x})$ in pre-chosen network layers and predictions $h_{y^*}(\mathbf{x})$, $h_{y^c}(\mathbf{x})$, $s(\mathbf{x})$. The attributions of $y^*$, $y^c$ and $s(\mathbf{x})$ to $\mathbf{x}$, i.e. $\mathcal{A}(\mathbf{x}, y^*)$, $\mathcal{A}(\mathbf{x}, y^c)$, $\mathcal{A}(\mathbf{x}, s(\mathbf{x}))$ are then computed with~(\ref{eq:gradient}), which reduce to a backpropagation step with respect to the desired layer activations and a few additional operations. Finally, the three attributions are combined with (\ref{equ:self_dis}). Thresholding the resulting heatmap with (\ref{eq:segdisc}) produces the discriminant explanation $\mathbf{r}\{y^*,y^c\}({\mathbf x})$. To further obtain a counterfactual explanation, the network is simply applied to $\mathbf{x}^c$ and $\mathbf{r}\{y^c,y^*\}({\mathbf x}^c)$ computed.

% An attribution map $m^p_{i,j}$ measures the contribution of the activations
% ${\bf a}_{i,j}$ at location $(i,j)$ to prediction $p$. This could be a class prediction or a confidence prediction. Conceptually,
% there is no difference between the two, we simply denote
% $p = g_p({\bf A})$, where ${\bf g}$ is the mapping from activation
% tensor ${\bf A}$ into prediction
% vector ${\bf g}({\bf A}) \in [0,1]^P$. For class predictions $P=C$, the
% prediction is a class, and $g_y({\bf A}({\bf x}))=f_y({\bf x})$.
% For confidence predictions $P=1$, the prediction is a confidence
% score, and $g({\bf A}({\bf x}))=s({\bf x})$.
% In any case, our goal is not to propose a new attribution function, but to leverage them on counterfactual visual explanations.
% We emphasize that any of the previous gradient based methods can be used in our strategy
% They are all calculated by the activation and its gradient. Suppose the general mapping is $q(\cdot)$, we currently use the

\section{Evaluation}

\noindent{\bf Challenges:} Explanations are difficult to evaluate
because ground truth is unavailable. Previous works
mainly presented qualitative
results~\cite{hendricks2018generating,pmlr-v97-goyal19a}.
\cite{pmlr-v97-goyal19a} also performed a human evaluation on MTurk,
using a machine teaching task. However, this evaluation had a few flaws,
which are discussed in Section \ref{sec:app_mt}. In any case, human
evaluation is cumbersome and difficult to replicate.
To avoid this, we introduce an alternative evaluation strategy based on the
proxy task of localization. Because this leverages datasets with annotations
for part locations and attributes\footnote{note that
  part and attribute annotations are only required for performance
  evaluation, not to compute the visualizations.}, we sometimes refer to image regions
(segments or keypoints) as parts.

\noindent{\bf Ground-truth:} The goal of counterfactual
explanations is to localize a region predictive of class $A$ but unpredictive
of class $B$. Hence, parts with attributes specific to $A$ and that do not
appear in $B$ can be seen as ground truth counterfactual regions.
This enables the evaluation of counterfactual
explanations as a part localization problem.
%The idea is similar to
%perturbation based counterfactual
%visualizations~\cite{dhurandhar2018explanations} where the class-sensitive
%regions of the query image are optimized and localized.
To synthesize ground truth, the $k^{th}$ part of an object of class $c$ is
represented by a semantic descriptor $\phi^k_c$ containing the attributes
present in this class. For example, an ``eye'' part can have color attributes ``red'', ``blue'', ``grey'', etc. The descriptor is a probability distribution over these attributes, characterizing the attribute variability of the part under each class.

The dissimilarity between classes $a$ and $b$, according to part $k$, is defined as
$\alpha^k_{a,b} = \gamma(\phi^k_a, \phi^k_b)$, where $\gamma(.,.)$ is
a dataset dependent function. Large dissimilarities indicate
that part $k$ is a discriminant for classes $a$ and $b$.
The values of $\alpha^k_{a,b}$ are computed for all class pairs $(a,b)$
and parts $\mathbf{p}_k$. The $M$ triplets
$\mathcal{G} = \{(\mathbf{p}_i, a_i, b_i)\}^M_{i=1}$ of largest dissimilarity
are selected as counterfactual ground-truth.

\noindent{\bf Evaluation metrics:} The metrics of explanation performance depend on the nature of part annotations.
On datasets where part locations are labelled with a single point,
i.e. $\mathbf{p}_i$ is a point (usually the geometric center of the part),
the quality of region $\mathbf{r}\{a, b\}(\mathbf{x})$
is calculated by precision (P)
and recall (R), where $P = \frac{J}{|\{ k|\mathbf{p}_k \in \mathbf{r}\}|}$,
$R = \frac{J}{|\{ i|(\mathbf{p}_i, a_i, b_i) \in \mathcal{G}, a_i = a, b_i = b\}|}$, and
$J = |\{i|\mathbf{p}_i \in \mathbf{r}, a_i = a, b_i = b\}|$ is the number of
included ground truth parts of generated regions.
%In order to consider both the recall and precision simultaneously, another metric $F_1$ score is used, which is the harmonic mean of the precision and recall, where an $F_1$ score reaches its best value at $1$ (perfect precision and recall) and worst at $0$.
Precision-recall curves are produced by varying the threshold $T$ used in
(\ref{eq:segdisc}). For datasets where parts are annotated with segmentation
masks, the quality of $\mathbf{r}\{a, b\}({\bf x})$ is evaluated using the
intersection over union (IoU) metric $\text{IoU} = \frac{|\mathbf{r} \cap \mathbf{p}|}
{|\mathbf{r} \cup \mathbf{p}|}$, where $\mathbf{p} = \{ \mathbf{p}_i|(\mathbf{p}_i, a_i, b_i) \in \mathcal{G}, a_i = a, b_i = b\}$.

For counterfactual explanations, we define a measure of the semantic
consistency of two segments, $\mathbf{r}\{a, b\}(\mathbf{x})$ and $\mathbf{r}\{b, a\}(\mathbf{x}^c)$, by calculating the
consistency of the parts included in them. This is denoted as
the part IoU (PIoU),
\begin{equation}
  \text{PIoU} = {\scriptstyle \frac{|\{k| (\mathbf{p}_k, a, b) \in \mathbf{r}\{a,b\}({\bf x})\}
  \cap \{ k|(\mathbf{p}_k, b, a) \in \mathbf{r}\{b,a\}({\bf x}^c)\}|}
{|\{ k|(\mathbf{p}_k, a, b) \in \mathbf{r}\{a,b\}({\bf x}) \}
  \cup \{ k| (\mathbf{p}_k, b, a) \in \mathbf{r}\{b,a\}({\bf x}^c)
\}|} }.
  \label{eq:partIoU}
\end{equation}

%\begin{figure*}[t]
%  \centering
%  \setlength{\tabcolsep}{2pt}
%  \begin{tabular}{cccc}
%    \includegraphics[width=.25\linewidth]{figures/precision_recall_curve_random_unique_a_as_std.pdf}
%    \includegraphics[width=.25\linewidth]{figures/precision_recall_curve_random_a_ab_abs_std.pdf}
%    \includegraphics[width=.25\linewidth]{figures/precision_recall_curve_alex_unique_a_as_std.pdf}
%    \includegraphics[width=.25\linewidth]{figures/precision_recall_curve_alex_a_ab_abs_std.pdf}
%  \end{tabular}
%  \caption{Impact of each component of~(\ref{equ:self_dis}) on precision-recall
%    (CUB200): Left two: beginners, right two: advanced users.}
%  \label{fig:CUB1}
%\end{figure*}

These metrics allow the quantitative comparison of different counterfactual explanation methods. On datasets with point-based ground truth, this is based on precision and recall of the generated counterfactual regions. On datasets with mask-based ground truth, the IoU is used. After conducting the whole process on both $\mathbf{x}$ and $\mathbf{x}^c$, PIoU can be computed to further measure the semantic matching between the highlighted regions in the two images. As long as the compared counterfactual regions of different methods have the same size, the comparison is fair. For SCOUT, region size can be controlled by manipulating
$T$ in (\ref{eq:segdisc}) and (\ref{eq:segcounter}) .

User expertise has an impact on counterfactual explanations. Beginner users tend to choose random counterfactual classes, while experts tend to pick counterfactual classes similar to the true class. Hence, explanation performance should be measured over the two user types. In this paper, users are simulated by choosing a random counterfactual class $b$  for beginners  and the class predicted by a small CNN for advanced users. Class $a$ is the prediction of the classifier used to generate the explanation, which is a larger CNN.

\section{Experiments}

%There is few suitably annotated fine-grained dataset available for our task. Most of fine-grained datasets like CARS196~\cite{KrauseStarkDengFei-Fei_3DRR2013}, Dogs~\cite{KhoslaYaoJayadevaprakashFeiFei_FGVC2011} only have image-level category label and do not provide localizable attributes annotations.  Other datasets like AWA~\cite{lampert2009learning}, SUN attribute~\cite{patterson2014sun} have attribute annotation, but they can not be positioned. Although the explanations could be generated, quantitative evaluations are undoable. OIDaircraft~\cite{mahendran14understanding} has $5$ part location annotation for airplanes, but the object category is given by airline which are not consistent with human intuitive perception for airplane classification like as production age or usage function. So this dataset has rarely been used in passed years. The lack of usable datasets in turn imply that fine-grained domain annotation is nontrivial and counterfactual explanation is mostly needed for teaching human labeling.
All experiments are performed on two datasets. CUB200~\cite{WelinderEtal2010} consists of $200$ fine-grained bird classes, annotated with $15$ part locations (points) including back, beak, belly, breast, crown, forehead, left/right eye, left/right leg, left/right wing, nape, tail and throat. Each part is associated with attribute information~\cite{WelinderEtal2010} and dissimilarities $\alpha^k_{a,b}$ are computed with
$\gamma(\phi^k_a,\phi^k_b) = e^{\{\text{KL}(\bm{\phi}^k_a || \bm{\phi}^k_b) + \text{KL}(\bm{\phi}^k_b || \bm{\phi}^k_a)\}}$~\cite{endres2003new}, where $\phi^k_c$ is a probability distribution over all attributes of the $k^{th}$ part under class $c$ and $\text{KL}$ is the Kullback-Leibler divergence. $M$ is chosen to leave $80\%$ largest triplets $({\bf p}_i,a_i,b_i)$ as ground truth. The majority of $({\bf p}_i,a_i,b_i)$ are selected because dissimilar parts dominate in $\alpha^k_{a,b}$ space.

\begin{figure}[t]
  \centering
  \setlength{\tabcolsep}{2pt}
  \begin{tabular}{cccc}
    \includegraphics[width=.5\linewidth]{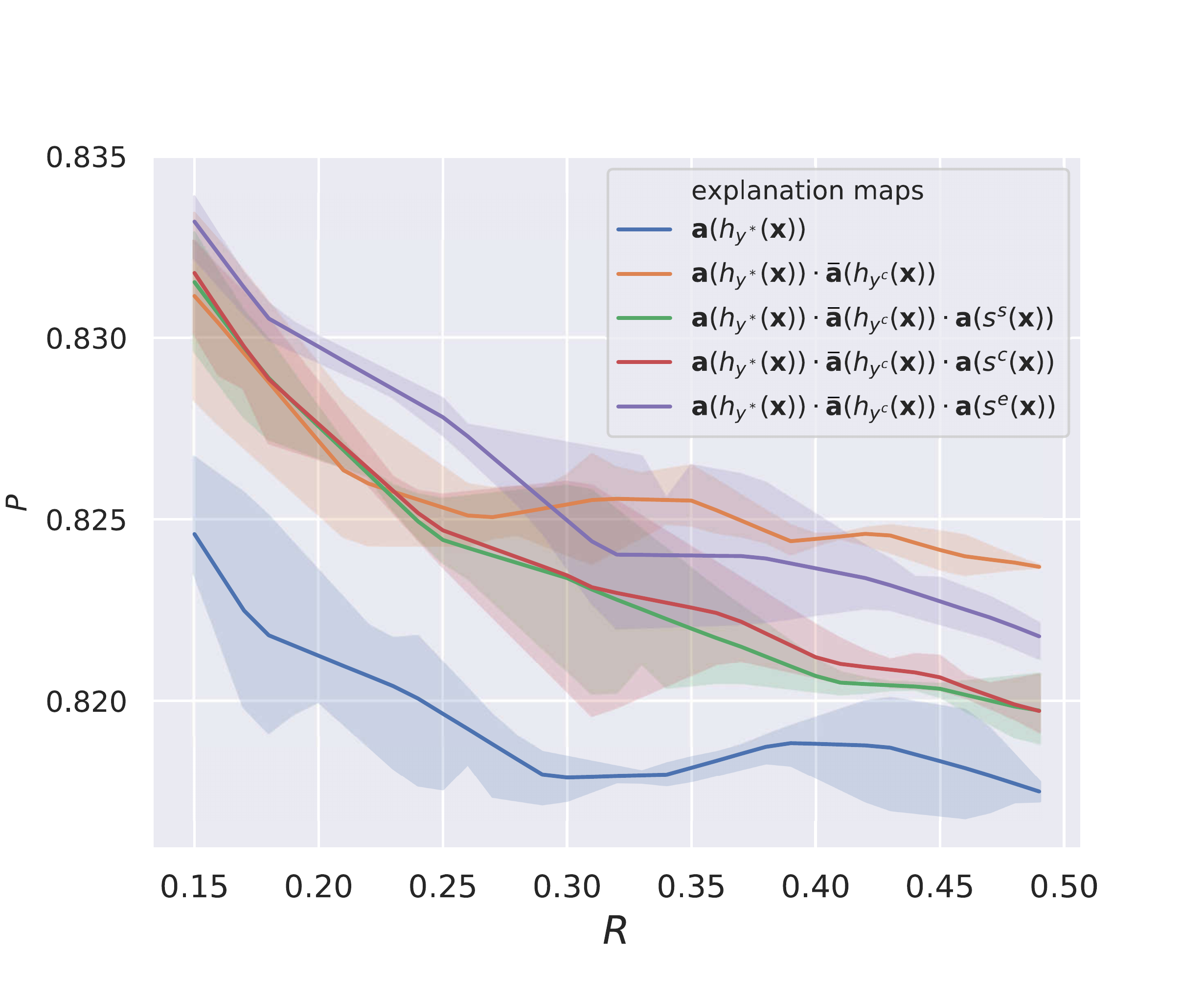}
    \includegraphics[width=.5\linewidth]{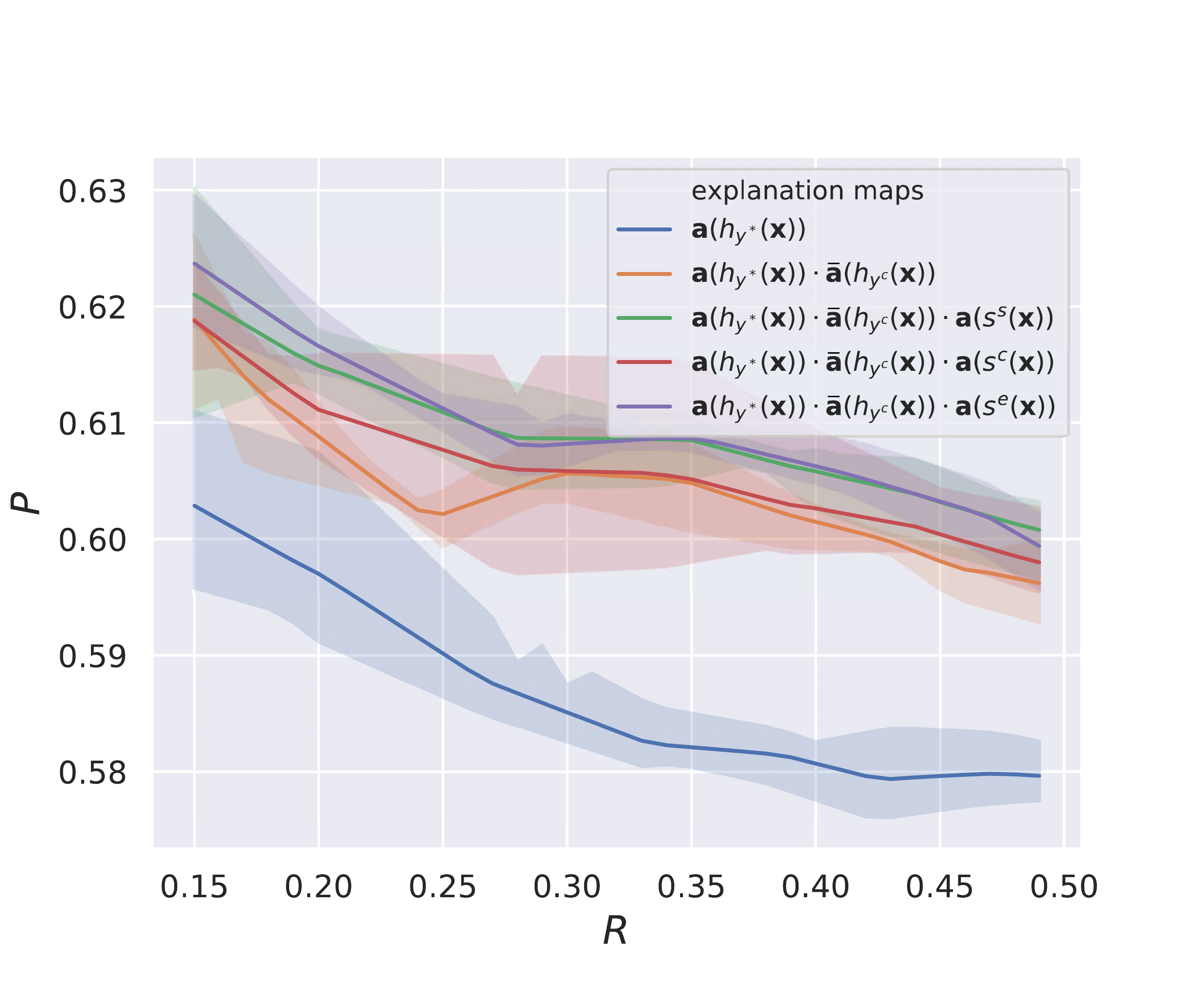}
  \end{tabular}
  \caption{Comparison to attributive explanations.
    Left: beginners, right: advanced users.}
  \label{fig:CUB1}
\end{figure}

The second dataset is ADE20K~\cite{zhou2017scene} with more than $1000$ fine-grained scene categories.
Segmentation masks are given for $150$ objects. In this case, objects are seen as scene parts and each object has a single attribute, i.e. $\phi^k_c$ is scalar (where $k \in \{1,..., 150\}$), which is the probability of occurrence of the object in a scene of class $c$. This is estimated by the relative frequency with which the part appears in scenes of class $c$. Ground truth consists of the triplets $({\bf p}_i,a_i,b_i)$ with $\phi^k_a > 0$ and $\phi^k_b = 0$, i.e. where object $k$ appears in class $a$ but not in class $b$.

In the discussion below, results are obtained on CUB200, except as otherwise stated.
ADE20K results are presented in the supplementary
materials. Unless otherwise
noted, visualizations are based on the last convolutional layer output of VGG16~\cite{simonyan2014very}, a
widely used network in visualization papers.
All counterfactual explanation results are presented for two types of virtual users. Randomly chosen labels mimic beginners while AlexNet predictions~\cite{krizhevsky2009learning} mimic advanced users.

%\subsection{Ablation study}

% We run several ablations to analyze the proposed counterfactual
% explanation algorithm. Since the results of the two datasets
% were qualitatively similar,

%\textbf{Confidence scores:} Figure \ref{fig:CUB1} show that, although all confidence scores improve performance, the gains are much larger for the easiness score. This is used in all remaining experiments.

%{\color{red} \textbf{Attribution maps and Architecture:} Please see supplemental materials.}

%\begin{figure}[t]
%  \centering
%  \setlength{\tabcolsep}{2pt}
%  \begin{tabular}{ccc}
%    \includegraphics[width=.5\linewidth]{figures/precision_recall_curve_contributionmapComp_std.pdf}
%    &\includegraphics[width=.5\linewidth]{figures/precision_recall_curve_architectureComp_std.pdf}
%  \end{tabular}
%  \caption{Impact of attribution function (left) and architecture (right)
%    on precision-recall (CUB200).}
%  \label{fig:CUB2}
%\end{figure}

\subsection{Comparison to attributive explanations}

Figure \ref{fig:CUB1} compares the discriminant explanations of
(\ref{equ:self_dis}), to attributive explanations
$\mathbf{a}(h_{y^*}(\mathbf{x}))$,
%$\mathbf{a}(h_{y^c}(\mathbf{x}))$, $\mathbf{a}(s(\mathbf{x}))$, of
for the two user types. Several conclusions are
possible: 1) discriminant maps significantly outperform attributions
for both user types, independently of the
confidence score used; 2) best performance is achieved with the
easiness score of~(\ref{eq:ea}); 3) the gains are larger
for expert users than beginners.
This is because the counter and predicted classes tend to be more similar for
the former and the corresponding attribution maps overlap.
In this case, pure attributive explanations are very uninformative.
The result also shows that self-awareness is most useful in expert domains.

\subsection{Comparison to state of the art}

\begin{table}
\setlength{\abovecaptionskip}{-2.0pt}
\setlength{\tabcolsep}{2pt}
\footnotesize
\begin{center}
\begin{tabular}{|c|c|c|c||c|c|}
\hline
\multicolumn{2}{|c|}{}& \multicolumn{2}{|c||}{Beginner User} & \multicolumn{2}{|c|}{Advanced User} \\
\hline
  Arch. &Metric & Goyal~\cite{pmlr-v97-goyal19a}& SCOUT & Goyal~\cite{pmlr-v97-goyal19a}& SCOUT \\
  \hline
\multirow{4}{*}{VGG16} & R & 0.02 (0.01)& {\bf 0.05} (0.01) & {\bf 0.05} (0.00)&{\bf 0.05}  (0.00) \\
& P & 0.76 (0.01) & {\bf 0.84} (0.01) & 0.56 (0.01) &{\bf 0.64} (0.01)\\
& PIoU & 0.13 (0.00) & {\bf 0.15} (0.00) &  0.09 (0.00) &{\bf 0.14} (0.02)\\
& IPS & 0.02 (0.00) &{\bf 26.51} (0.71) & & \\
\hline
\multirow{4}{*}{ResNet-50} & R & 0.03 (0.01) & {\bf 0.09} (0.02) & 0.12 (0.01) & {\bf 0.16} (0.00)  \\
& P & 0.77 (0.01) & {\bf 0.81} (0.01) & 0.57  (0.02) & {\bf 0.60} (0.01)\\
& PIoU &{\bf 0.18}  (0.01) & 0.16 (0.01) &{\bf 0.15}  (0.00) &{\bf 0.15} (0.01)\\
& IPS & 1.13 (0.07) &{\bf 78.54} (11.87)& & \\
  \hline
\end{tabular}
\end{center}
\caption{Comparison to the state of the art. (IPS: images per second, implemented on NVIDIA TITAN Xp. Results are shown as mean(stddev))}
\label{tab:comp}
\end{table}

Table \ref{tab:comp} presents a comparison between SCOUT and the method of \cite{pmlr-v97-goyal19a} which obtained the best results by exhaustive search, for
the two user types. For fair comparison, these experiments use the softmax
score of (\ref{eq:so}), so that model sizes are equal for both approaches.
The size of the counterfactual region is the receptive field size of one
unit ($\frac{1}{14*14} \approx 0.005$ of image size
on VGG16 and $\frac{1}{7*7} \approx 0.02$ on ResNet-50).
This was constrained by the speed of the algorithm of \cite{pmlr-v97-goyal19a},
where the counterfactual region is detected by exhaustive feature matching.

\begin{figure}[t]
  \centering
  \setlength{\tabcolsep}{2pt}
  \begin{tabular}{ccc}
    \includegraphics[width=.5\linewidth]{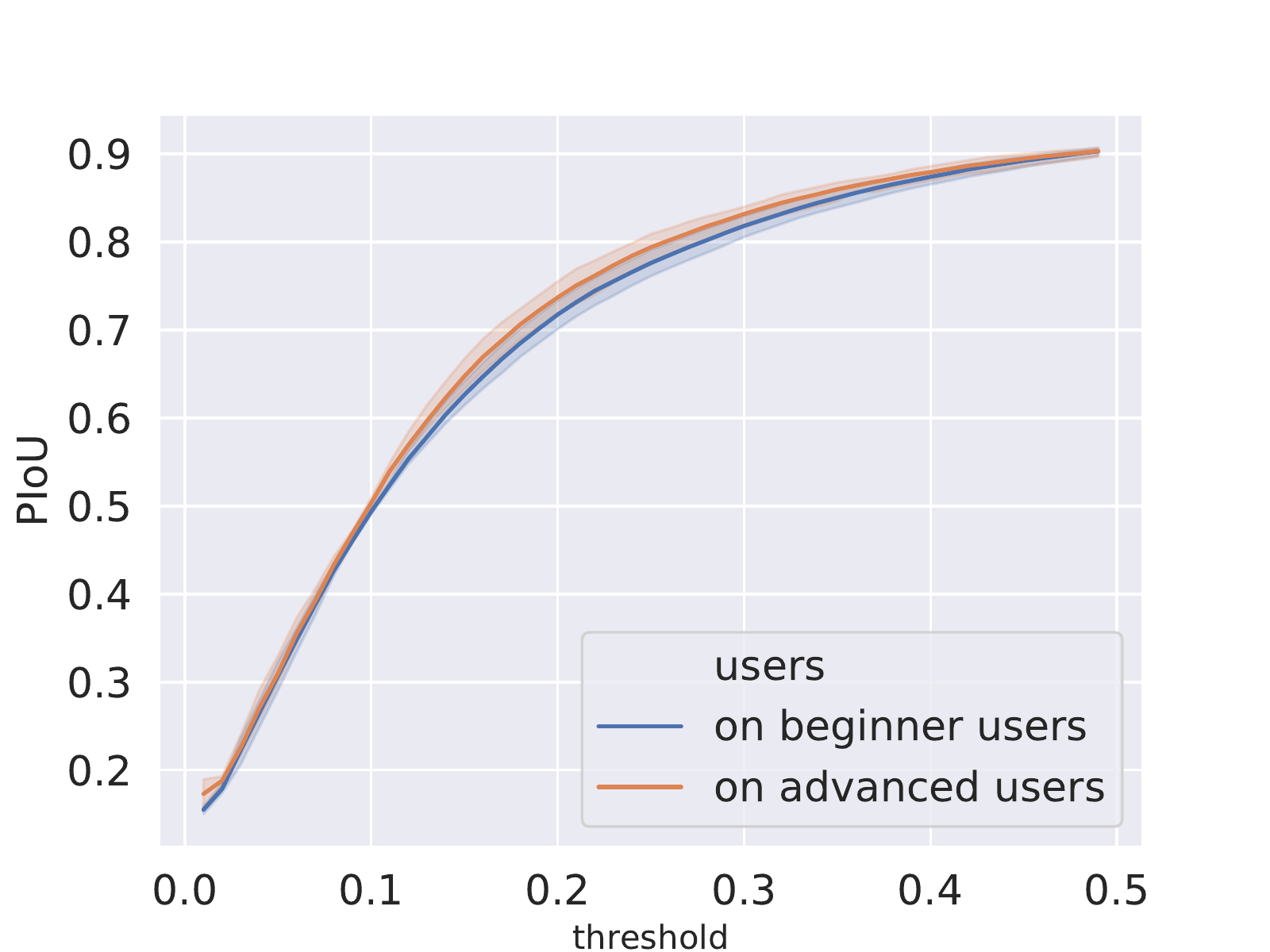}
    &\includegraphics[width=.5\linewidth]{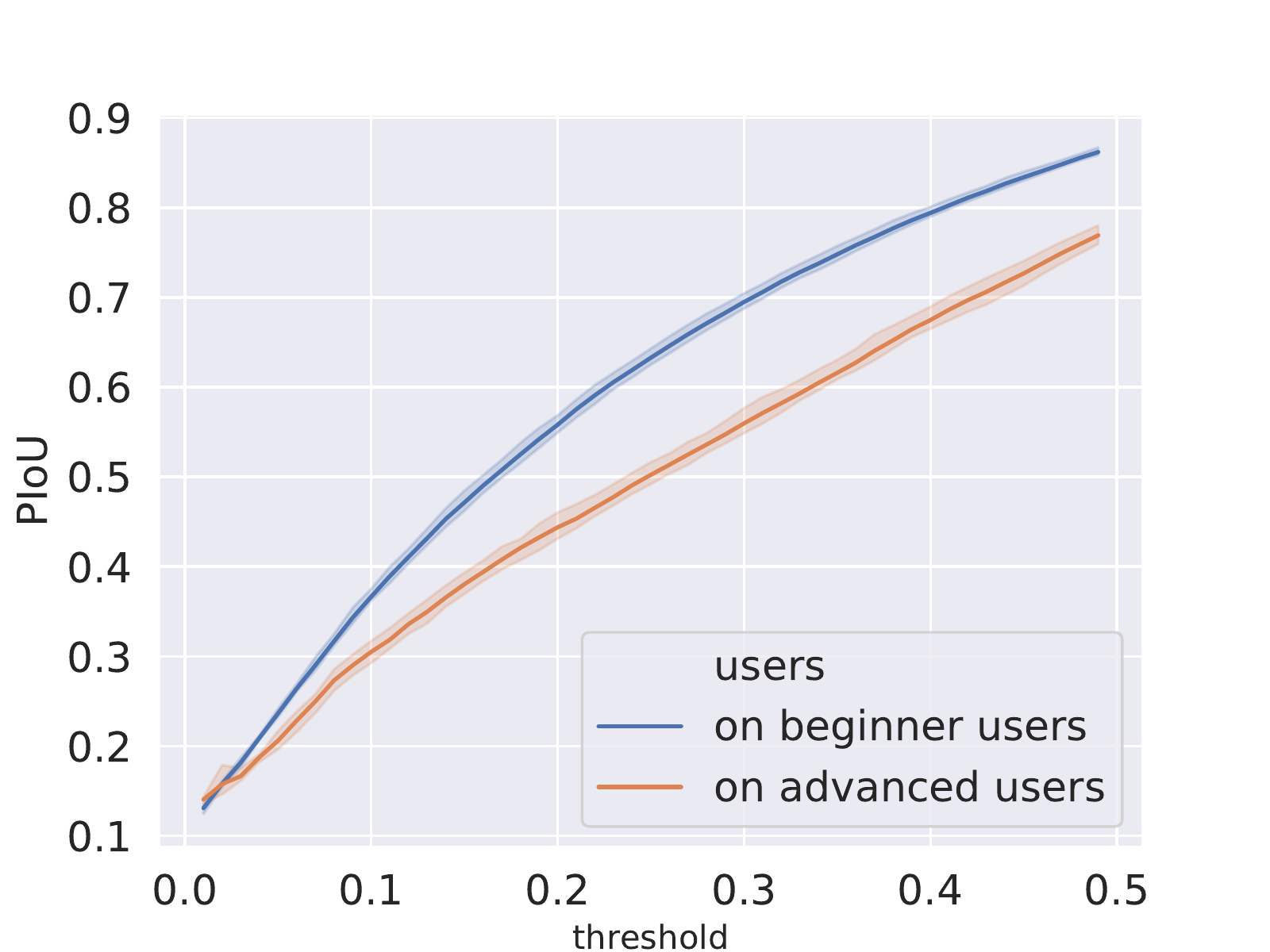}
  \end{tabular}
  \caption{PIoU of SCOUT as a function of the segmentation threshold on CUB200.
    Left:  VGG16, right: ResNet-50.}
  \label{fig:pointIoU_thresholds}
\end{figure}

Several conclusions could be drawn from the table. First, SCOUT outperforms \cite{pmlr-v97-goyal19a} in almost all cases.
Second, SCOUT is much faster, improving the speed
of \cite{pmlr-v97-goyal19a} by $1000+$ times on VGG and $50+$ times on ResNet.
This is because it does not require exhaustive feature matching.
These gains increase with the size of the counterfactual region, since
computation time is constant for the proposed approach but exponential on
region size for \cite{pmlr-v97-goyal19a}.
Third, due to the small size used in these experiments,
PIoU is relatively low for both methods. It is, however, larger for
the proposed explanations with large gains in some cases (VGG $\&$ advanced).
Figure \ref{fig:pointIoU_thresholds} shows that the PIoU can raise up to
$0.5$ for regions of $10\%$ image size (VGG) or $20\%$ (ResNet).
This suggests that, for regions of this size, the region pairs
have matching semantics.

\subsection{Visualizations}

\begin{figure}[t]
  \centering
  \includegraphics[width=0.49\textwidth]{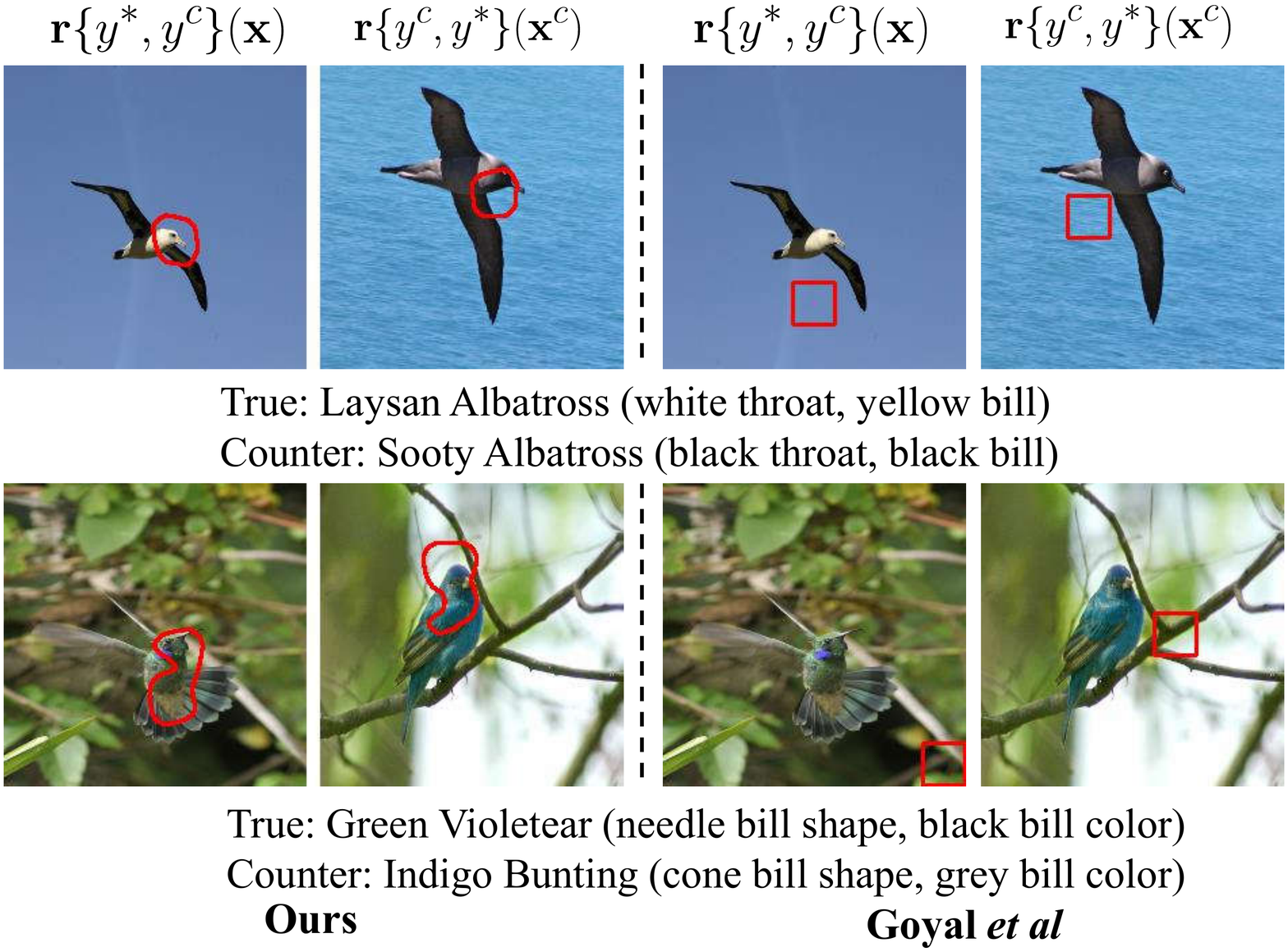}
  \caption{Comparison of counterfactual explanations (true and counter
    classes shown below each example, and ground truth class-specific part
    attributes in parenthesis).}
  \label{fig:kp_comp}
\end{figure}

\begin{figure}[t]
  \centering
  \includegraphics[width=0.49\textwidth]{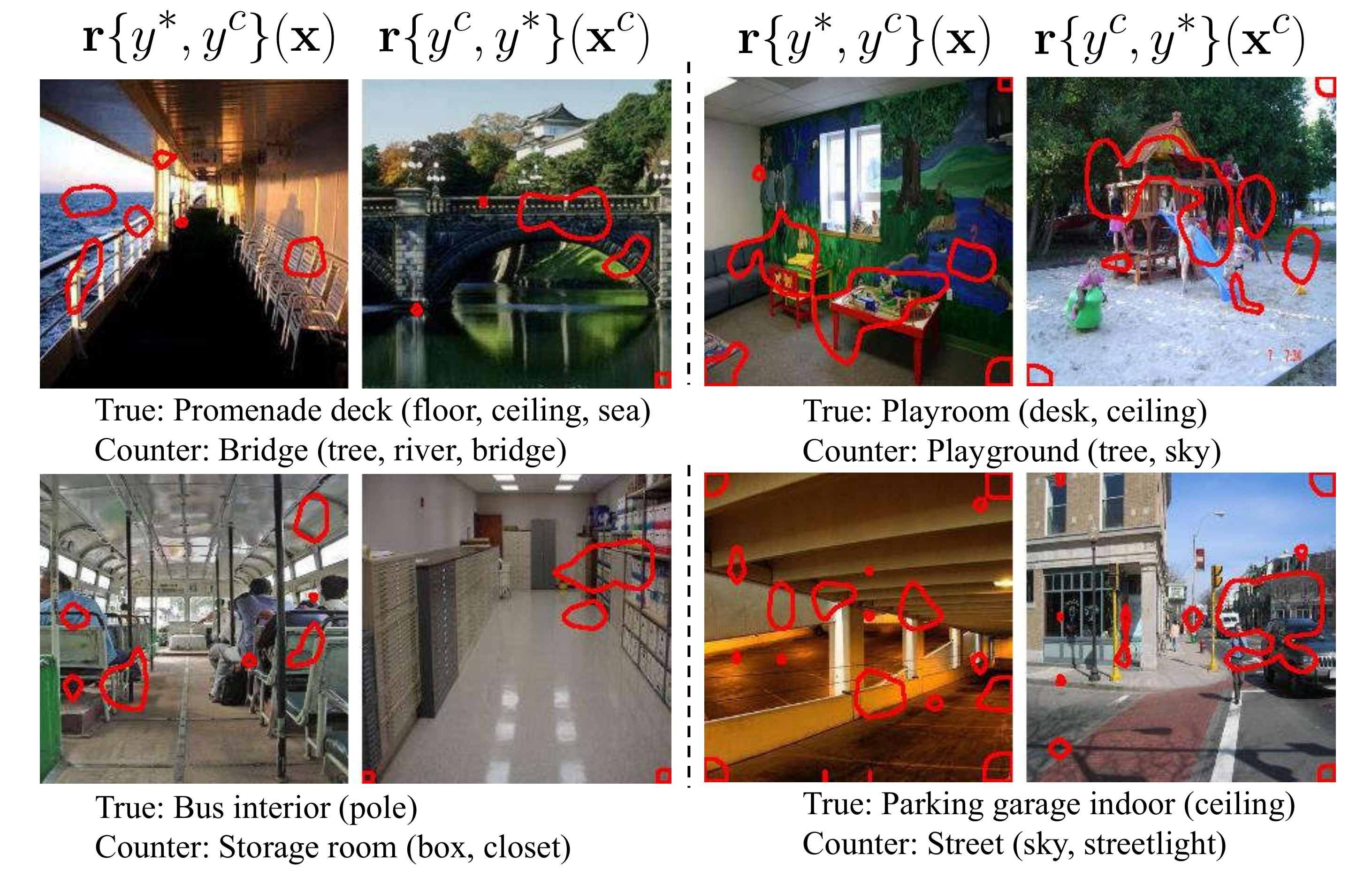}
  \caption{Counterfactual explanations on ADE20K.}
  \label{fig:seg_matching}
\end{figure}

Figure \ref{fig:kp_comp} shows two examples of counterfactual visualizations derived from the ResNet50
on CUB200. The regions selected in the query and counter class image are
shown in red. The true $y^*$ and counter $y^c$ class are shown below the
images and followed by the ground truth discriminative attributes for the
image pair. Note how the proposed explanations identify semantically matched
and class-specific bird parts on both images. For example, the throat and bill
that distinguish Laysan from Sooty Albatrosses. This feedback enables a
user to learn that Laysans have white throats and yellow bills, while
Sootys have black throats and bills.
% So if a learner ask ``Why it is not a Sooty Albatross?'' Our explanation would be ``Because the highlighting region is different from Sooty Albatross, please refer to the accompanied Laysan Albatross image to see what Sooty Albatross should look like.'' The paired regions can give hint for learners to differentiate the two birds.
This is unlike the regions produced by \cite{pmlr-v97-goyal19a}, which
sometimes highlight irrelevant cues, such as the background.
Figure \ref{fig:seg_matching} presents similar figures for ADE20K, where the
proposed explanations tend to identify scene-discriminative objects.
For example, that a  promenade deck contains objects `floor', `ceiling',
`sea,' while a bridge scene includes `tree', `river' and `bridge'.

%\begin{figure*}[t]
%  \centering
%  \includegraphics[width=0.8\textwidth]{figures/keypoint_matching.pdf}
%  \caption{Keypoint matching}
%  \label{fig:keypoint_matching}
%\end{figure*}

%\begin{figure}[t]
%  \centering
%  \includegraphics[width=1.0\textwidth]{figures/pointIoU_threshold_users_std.pdf}
%  \caption{pointIoU under thresholds on CUB200}
%  \label{fig:pointIoU_thresholds}
%\end{figure}

% The parallel ablation study is also conducted on ADE20K, from which the same conclusion could be made. Due to the space limitation, results are given in the supplement (Table \ref{tab:ade_comp}) ({\color{red} I plan to move this huge table to supplementary materials.}).

%\begin{figure}[t]
%  \centering
%  \includegraphics[width=1.0\textwidth]{figures/pointIoU_threshold_users_std.pdf}
%  \caption{pointIoU under thresholds on CUB200}
%  \label{fig:pointIoU_thresholds}
%\end{figure}

%\begin{figure}[t]
%  \begin{minipage}{.5\linewidth}
%    \includegraphics[width=\linewidth]{figures/pointIoU_threshold_users_std.pdf}
%    \caption{pointIoU under thresh}
%    \label{fig:pointIoU}
%  \end{minipage}
%  \begin{minipage}{.5\linewidth}
%    \includegraphics[width=\linewidth]{figures/pointIoU_threshold_users_std.pdf}
%    \caption{pointIoU under thresholds on CUB200}
%    \label{fig:pointIoU_thresholds}
%  \end{minipage}
%\end{figure}

\subsection{Application to machine teaching}
\label{sec:app_mt}

% \begin{figure}[t]
%   \centering
%   \setlength{\tabcolsep}{2pt}
%   \begin{tabular}{ccc}
%     \includegraphics[width=0.5\linewidth]{figures/interface_icml19.pdf}
%     &\includegraphics[width=0.5\linewidth]{figures/interface_ours.pdf}
%   \end{tabular}
%   \caption{Comparison of two feedback interfaces with an example of Kentucky Warbler and Setophaga Citrina (left column: query images; right column: distractor images.}
%   \label{fig:interface}
% \end{figure}

\begin{figure}
\centering
% \begin{subfigure}[t]{0.45\textwidth}
%   \includegraphics[width=1.0\linewidth]{figures/interface_icml19.pdf}
  % \caption{Goyal at al~\cite{pmlr-v97-goyal19a}}
%  \label{fig:interface_icml}
%\end{subfigure}
%\begin{subfigure}[t]{0.45\textwidth}
\includegraphics[width=.8\linewidth]{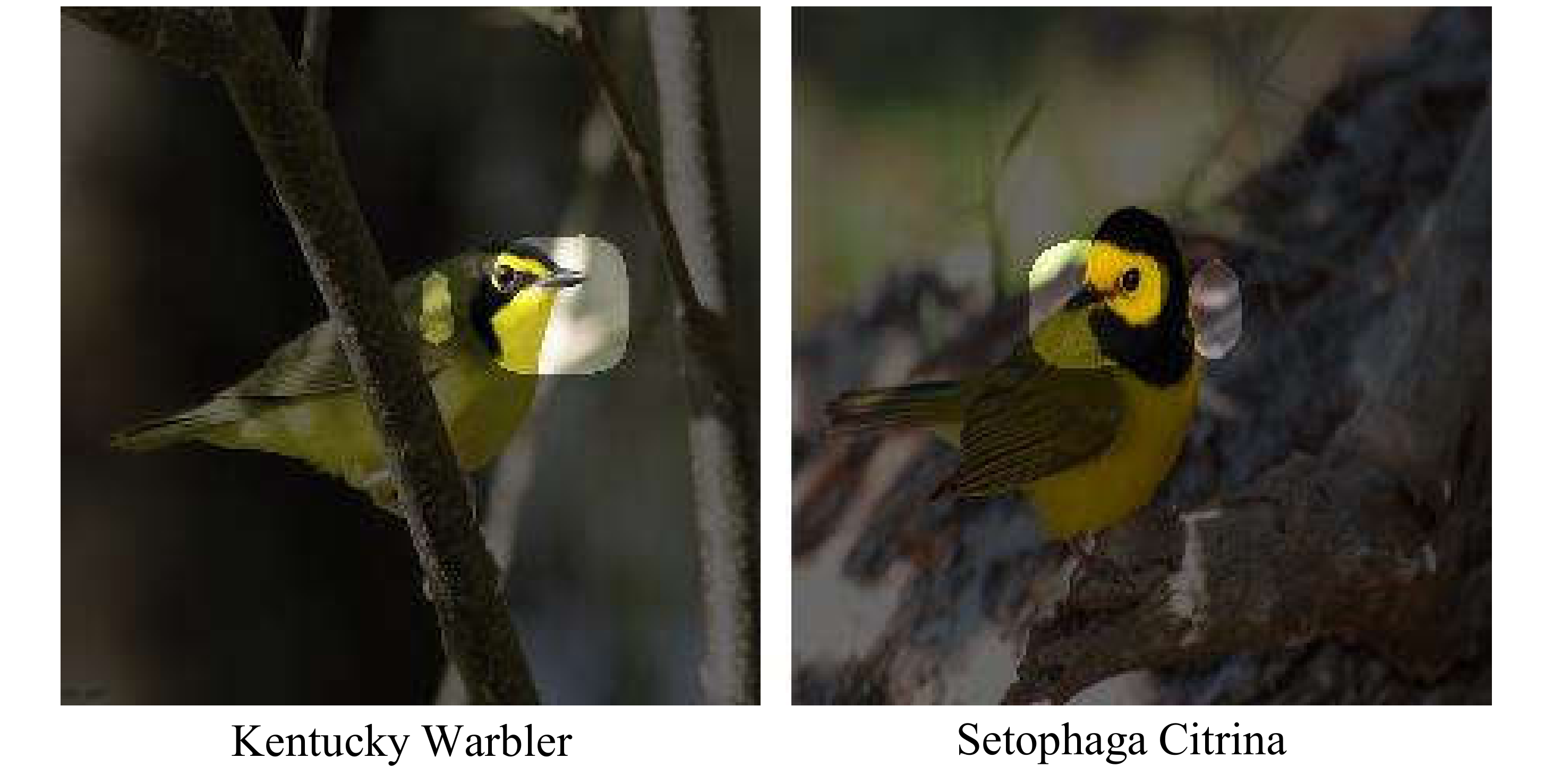}
%  \caption{Ours}
%\end{subfigure}
\caption{Visualization of machine teaching experiment.}
\label{fig:interface_ours}
\end{figure}

\cite{pmlr-v97-goyal19a} used counterfactual explanations to design
an experiment to teach humans distinguish two bird classes.
During a training stage, learners are asked to classify birds. When they
make a mistake, they are shown counterfactual feedback of the type of
Figure~\ref{fig:kp_comp}, using the true class as $y^*$ and the class
they chose as $y^c$.
% This is of the form the image shown is not of a Bravo (as you suggest) but of an Alpha. However, if the image region highlighted in the left looked like the highlighted in the right, then the image would be of a Bravo.
This helps them understand why they chose the wrong label,
and learn how to better distinguish the classes. In a test stage,
learners are then asked to classify a bird without visual aids.
Experiments reported
in~\cite{pmlr-v97-goyal19a} show that this is much more effective than
simply telling them whether their answer is correct/incorrect,
or other simple training strategies.
We made two modifications to this set-up. The first was to replace bounding
boxes with highlighting of the counterfactual reasons, as shown in
Figure \ref{fig:interface_ours}. We also instructed
learners not to be distracted by the darkened regions. Unlike the set-up
of~\cite{pmlr-v97-goyal19a}, this guarantees
that they do not exploit cues outside the counterfactual regions
to learn bird differences.
% Second, to check this,
% we add a contrast experiment where highlighted regions are generated
% randomly (without telling the learners). If this produces the same
% results, one can conclude that the explanations do not promote
% learning.
Second, to check this,
we added two contrast experiments where 1) highlighted regions are generated
randomly (without telling the learners); 2) the entire images are lighted.
If these produce the same results, one can conclude that the explanations
do not promote learning.

We also chose two more difficult birds, the Setophaga Citrina and the
Kentucky Warbler (see Figure \ref{fig:interface_ours}), than those used in
\cite{pmlr-v97-goyal19a}. This is because these classes have large
intra-class diversity. The two classes also  cannot be distinguished by color
alone, unlike those used in \cite{pmlr-v97-goyal19a}.
%The hard task can better evaluate different counterfactual visual explanations in machine teaching.
%Graduate students with no particular expertise in biology or zoology were
%recruited for the experiments because of the level of difficulty of the task.
The experiment has three steps. The first is a pre-learning test,
where students are asked to classify $20$ examples of the two classes,
or choose a `Don't know' option. The second is a learning stage,
where counterfactual explanations are provided for $10$ bird pairs.
The third is a post-learning test, where students are asked to answer
$20$ binary classification questions. In this experiment, all students chose `Don't know' in the
pre-learning test.
However, after the learning step, they achieved $95\%$ mean accuracy,
compared to $60\%$ (random highlighted
regions) and $77\%$ (entire images lighted) in the contrast settings. These results suggest that SCOUT can help teach non-expert humans distinguish categories
from an expert domain.

\section{Conclusion}

In this work, we proposed a new family of discriminant
explanations, which leverage self-awareness and bridge the gap between
attributions and counterfactuals. A quantitative evaluation
protocol was also proposed. Experiments under both this protocol and
machine teaching experiments show that both the proposed discriminant
and counterfactual explanations achieve much better
performance than existing attributive and counterfactual methods.
\paragraph{Acknowledgements} This work was partially funded by NSF awards IIS-1637941, IIS-1924937, and NVIDIA GPU donations.

{\small
\bibliographystyle{ieee_fullname}
\bibliography{egbib}

\begin{thebibliography}{10}\itemsep=-1pt

\bibitem{ancona2017unified}
Marco Ancona, Enea Ceolini, Cengiz {\"O}ztireli, and Markus Gross.
\newblock A unified view of gradient-based attribution methods for deep neural
  networks.
\newblock In {\em NIPS 2017-Workshop on Interpreting, Explaining and
  Visualizing Deep Learning}. ETH Zurich, 2017.

\bibitem{Hendricks_2018_ECCV}
Lisa Anne~Hendricks, Ronghang Hu, Trevor Darrell, and Zeynep Akata.
\newblock Grounding visual explanations.
\newblock In {\em The European Conference on Computer Vision (ECCV)}, September
  2018.

\bibitem{bach2015pixel}
Sebastian Bach, Alexander Binder, Gr{\'e}goire Montavon, Frederick Klauschen,
  Klaus-Robert M{\"u}ller, and Wojciech Samek.
\newblock On pixel-wise explanations for non-linear classifier decisions by
  layer-wise relevance propagation.
\newblock {\em PloS one}, 10(7):e0130140, 2015.

\bibitem{basu2013teaching}
Sumit Basu and Janara Christensen.
\newblock Teaching classification boundaries to humans.
\newblock In {\em Twenty-Seventh AAAI Conference on Artificial Intelligence},
  2013.

\bibitem{bendale2016towards}
Abhijit Bendale and Terrance~E Boult.
\newblock Towards open set deep networks.
\newblock In {\em Proceedings of the IEEE conference on computer vision and
  pattern recognition}, pages 1563--1572, 2016.

\bibitem{deruyttere2019talk2car}
Thierry Deruyttere, Simon Vandenhende, Dusan Grujicic, Luc Van~Gool, and
  Marie-Francine Moens.
\newblock Talk2car: Taking control of your self-driving car.
\newblock {\em arXiv preprint arXiv:1909.10838}, 2019.

\bibitem{devries2018learning}
Terrance DeVries and Graham~W Taylor.
\newblock Learning confidence for out-of-distribution detection in neural
  networks.
\newblock {\em arXiv preprint arXiv:1802.04865}, 2018.

\bibitem{dhurandhar2018explanations}
Amit Dhurandhar, Pin-Yu Chen, Ronny Luss, Chun-Chen Tu, Paishun Ting,
  Karthikeyan Shanmugam, and Payel Das.
\newblock Explanations based on the missing: Towards contrastive explanations
  with pertinent negatives.
\newblock In {\em Advances in Neural Information Processing Systems}, pages
  592--603, 2018.

\bibitem{endres2003new}
Dominik~Maria Endres and Johannes~E Schindelin.
\newblock A new metric for probability distributions.
\newblock {\em IEEE Transactions on Information theory}, 2003.

\bibitem{geifman2017selective}
Yonatan Geifman and Ran El-Yaniv.
\newblock Selective classification for deep neural networks.
\newblock In {\em Advances in neural information processing systems}, pages
  4878--4887, 2017.

\bibitem{goodfellow2014generative}
Ian Goodfellow, Jean Pouget-Abadie, Mehdi Mirza, Bing Xu, David Warde-Farley,
  Sherjil Ozair, Aaron Courville, and Yoshua Bengio.
\newblock Generative adversarial nets.
\newblock In {\em Advances in neural information processing systems}, pages
  2672--2680, 2014.

\bibitem{pmlr-v97-goyal19a}
Yash Goyal, Ziyan Wu, Jan Ernst, Dhruv Batra, Devi Parikh, and Stefan Lee.
\newblock Counterfactual visual explanations.
\newblock In Kamalika Chaudhuri and Ruslan Salakhutdinov, editors, {\em
  Proceedings of the 36th International Conference on Machine Learning},
  volume~97 of {\em Proceedings of Machine Learning Research}, pages
  2376--2384, 2019.

\bibitem{hendricks2018generating}
Lisa~Anne Hendricks, Ronghang Hu, Trevor Darrell, and Zeynep Akata.
\newblock Generating counterfactual explanations with natural language.
\newblock {\em arXiv preprint arXiv:1806.09809}, 2018.

\bibitem{hendrycks17baseline}
Dan Hendrycks and Kevin Gimpel.
\newblock A baseline for detecting misclassified and out-of-distribution
  examples in neural networks.
\newblock {\em Proceedings of International Conference on Learning
  Representations}, 2017.

\bibitem{hendrycks2018deep}
Dan Hendrycks, Mantas Mazeika, and Thomas~G Dietterich.
\newblock Deep anomaly detection with outlier exposure.
\newblock {\em Proceedings of International Conference on Learning
  Representations}, 2019.

\bibitem{krizhevsky2009learning}
Alex Krizhevsky and Geoffrey Hinton.
\newblock Learning multiple layers of features from tiny images.
\newblock Technical report, Citeseer, 2009.

\bibitem{le2004teaching}
Ronan Le~Hy, Anthony Arrigoni, Pierre Bessi{\`e}re, and Olivier Lebeltel.
\newblock Teaching bayesian behaviours to video game characters.
\newblock {\em Robotics and Autonomous Systems}, 47(2-3):177--185, 2004.

\bibitem{lee2017training}
Kimin Lee, Honglak Lee, Kibok Lee, and Jinwoo Shin.
\newblock Training confidence-calibrated classifiers for detecting
  out-of-distribution samples.
\newblock {\em Proceedings of International Conference on Learning
  Representations}, 2018.

\bibitem{li2020background}
Yi Li and Nuno Vasconcelos.
\newblock Background data resampling for outlier-aware classification.
\newblock In {\em Proceedings of the IEEE Conference on Computer Vision and
  Pattern Recognition}, 2020.

\bibitem{liang2017enhancing}
Shiyu Liang, Yixuan Li, and R Srikant.
\newblock Enhancing the reliability of out-of-distribution image detection in
  neural networks.
\newblock {\em Proceedings of International Conference on Learning
  Representations}, 2018.

\bibitem{liu2019generative}
Shusen Liu, Bhavya Kailkhura, Donald Loveland, and Yong Han.
\newblock Generative counterfactual introspection for explainable deep
  learning.
\newblock {\em arXiv preprint arXiv:1907.03077}, 2019.

\bibitem{liu2015faceattributes}
Ziwei Liu, Ping Luo, Xiaogang Wang, and Xiaoou Tang.
\newblock Deep learning face attributes in the wild.
\newblock In {\em Proceedings of International Conference on Computer Vision
  (ICCV)}, December 2015.

\bibitem{luss2019generating}
Ronny Luss, Pin-Yu Chen, Amit Dhurandhar, Prasanna Sattigeri, Karthikeyan
  Shanmugam, and Chun-Chen Tu.
\newblock Generating contrastive explanations with monotonic attribute
  functions.
\newblock {\em arXiv preprint arXiv:1905.12698}, 2019.

\bibitem{miller2018contrastive}
Tim Miller.
\newblock Contrastive explanation: A structural-model approach.
\newblock {\em arXiv preprint arXiv:1811.03163}, 2018.

\bibitem{piech2015deep}
Chris Piech, Jonathan Bassen, Jonathan Huang, Surya Ganguli, Mehran Sahami,
  Leonidas~J Guibas, and Jascha Sohl-Dickstein.
\newblock Deep knowledge tracing.
\newblock In {\em Advances in neural information processing systems}, pages
  505--513, 2015.

\bibitem{rathi2019generating}
Shubham Rathi.
\newblock Generating counterfactual and contrastive explanations using shap.
\newblock {\em arXiv preprint arXiv:1906.09293}, 2019.

\bibitem{scheirer2012toward}
Walter~J Scheirer, Anderson de Rezende~Rocha, Archana Sapkota, and Terrance~E
  Boult.
\newblock Toward open set recognition.
\newblock {\em IEEE transactions on pattern analysis and machine intelligence},
  35(7):1757--1772, 2012.

\bibitem{selvaraju2017grad}
Ramprasaath~R Selvaraju, Michael Cogswell, Abhishek Das, Ramakrishna Vedantam,
  Devi Parikh, and Dhruv Batra.
\newblock Grad-cam: Visual explanations from deep networks via gradient-based
  localization.
\newblock In {\em Proceedings of the IEEE International Conference on Computer
  Vision}, pages 618--626, 2017.

\bibitem{shrikumar2017learning}
Avanti Shrikumar, Peyton Greenside, and Anshul Kundaje.
\newblock Learning important features through propagating activation
  differences.
\newblock In {\em Proceedings of the 34th International Conference on Machine
  Learning-Volume 70}, pages 3145--3153. JMLR. org, 2017.

\bibitem{shrikumar2016not}
Avanti Shrikumar, Peyton Greenside, Anna Shcherbina, and Anshul Kundaje.
\newblock Not just a black box: Learning important features through propagating
  activation differences.
\newblock {\em arXiv preprint arXiv:1605.01713}, 2016.

\bibitem{simonyan2013deep}
Karen Simonyan, Andrea Vedaldi, and Andrew Zisserman.
\newblock Deep inside convolutional networks: Visualising image classification
  models and saliency maps.
\newblock {\em Workshop at International Conference on Learning
  Representations}, 2014.

\bibitem{simonyan2014very}
Karen Simonyan and Andrew Zisserman.
\newblock Very deep convolutional networks for large-scale image recognition.
\newblock {\em arXiv preprint arXiv:1409.1556}, 2014.

\bibitem{singla2014near}
Adish Singla, Ilija Bogunovic, G{\'a}bor Bart{\'o}k, Amin Karbasi, and Andreas
  Krause.
\newblock Near-optimally teaching the crowd to classify.
\newblock In {\em ICML}, page~3, 2014.

\bibitem{sundararajan2017axiomatic}
Mukund Sundararajan, Ankur Taly, and Qiqi Yan.
\newblock Axiomatic attribution for deep networks.
\newblock In {\em Proceedings of the 34th International Conference on Machine
  Learning-Volume 70}, pages 3319--3328. JMLR. org, 2017.

\bibitem{van2019interpretable}
Arnaud Van~Looveren and Janis Klaise.
\newblock Interpretable counterfactual explanations guided by prototypes.
\newblock {\em arXiv preprint arXiv:1907.02584}, 2019.

\bibitem{wachter2017counterfactual}
Sandra Wachter, Brent Mittelstadt, and Chris Russell.
\newblock Counterfactual explanations without opening the black box: Automated
  decisions and the gpdr.
\newblock {\em Harv. JL \& Tech.}, 31:841, 2017.

\bibitem{Wang_2018_ECCV}
Pei Wang and Nuno Vasconcelos.
\newblock Towards realistic predictors.
\newblock In {\em The European Conference on Computer Vision}, 2018.

\bibitem{pei2018deliberative}
Pei Wang and Nuno Vasconcelos.
\newblock Deliberative explanations: visualizing network insecurities.
\newblock In {\em Advances in Neural Information Processing Systems 32}, pages
  1374--1385, 2019.

\bibitem{wang2017idk}
Xin Wang, Yujia Luo, Daniel Crankshaw, Alexey Tumanov, Fisher Yu, and Joseph~E
  Gonzalez.
\newblock Idk cascades: Fast deep learning by learning not to overthink.
\newblock {\em arXiv preprint arXiv:1706.00885}, 2017.

\bibitem{WelinderEtal2010}
P. Welinder, S. Branson, T. Mita, C. Wah, F. Schroff, S. Belongie, and P.
  Perona.
\newblock {Caltech-UCSD Birds 200}.
\newblock Technical Report CNS-TR-2010-001, California Institute of Technology,
  2010.

\bibitem{zhou2016learning}
Bolei Zhou, Aditya Khosla, Agata Lapedriza, Aude Oliva, and Antonio Torralba.
\newblock Learning deep features for discriminative localization.
\newblock In {\em Proceedings of the IEEE conference on computer vision and
  pattern recognition}, pages 2921--2929, 2016.

\bibitem{zhou2017scene}
Bolei Zhou, Hang Zhao, Xavier Puig, Sanja Fidler, Adela Barriuso, and Antonio
  Torralba.
\newblock Scene parsing through ade20k dataset.
\newblock In {\em Proceedings of the IEEE Conference on Computer Vision and
  Pattern Recognition}, 2017.

\bibitem{zhou2018an}
Xiaojin Zhu, Adish Singla, Sandra Zilles, and Anna~N. Rafferty.
\newblock An overview of machine teaching.
\newblock {\em arXiv preprint arXiv:1801.05927}, 2018.

\bibitem{zintgraf2017visualizing}
Luisa~M Zintgraf, Taco~S Cohen, Tameem Adel, and Max Welling.
\newblock Visualizing deep neural network decisions: Prediction difference
  analysis.
\newblock {\em ICLR}, 2017.

\end{thebibliography}
}

\end{document}